\documentclass[letterpaper, 10 pt, conference]{ieeeconf}  

\IEEEoverridecommandlockouts                              

\usepackage{stfloats}
\usepackage{cite}
\usepackage{float}
\usepackage{bm}
\usepackage{here}
\usepackage{soul, color}
\usepackage{multirow}
\usepackage{tabularx}
\usepackage{algorithmic}
\usepackage{makecell}
\usepackage{hhline}
\usepackage{nicefrac}
\usepackage{lipsum}
\usepackage{array}
\usepackage{booktabs}
\usepackage{paralist}
\usepackage{gensymb}
\usepackage{inputenc}
\usepackage[dvipsnames]{xcolor}
\usepackage{colortbl}
\usepackage{svg}
\usepackage{threeparttable}
\usepackage{enumerate}
\usepackage{upgreek}
\usepackage[T1]{fontenc}
\usepackage{babel}
\usepackage{array}
\usepackage{textcomp}
\usepackage{mathtools}
\usepackage{amsmath}
\usepackage{amssymb}
\usepackage{graphicx}
\usepackage[unicode=true,
 bookmarks=true,bookmarksnumbered=true,bookmarksopen=true,bookmarksopenlevel=1,
 breaklinks=false,pdfborder={0 0 0},pdfborderstyle={},backref=false,colorlinks=false]
 {hyperref}
\hypersetup{pdftitle={Wirelessly-Controlled Untethered Piezoelectric Planar Soft Robot Capable of Bidirectional Crawling and Rotation},
 pdfauthor={Zhiwu Zheng, Hsin Cheng, Prakhar Kumar, Sigurd Wagner, Minjie Chen, Naveen Verma and James C. Sturm},
 pdfpagelayout=OneColumn, pdfnewwindow=true, pdfstartview=XYZ, plainpages=false}

\usepackage[caption=false]{subfig}

\usepackage{babel}

\title{\LARGE \bf
Wirelessly-Controlled Untethered Piezoelectric Planar Soft Robot Capable of Bidirectional Crawling and
Rotation}
\author{Zhiwu Zheng, Hsin Cheng, Prakhar Kumar, \\Sigurd Wagner, Minjie Chen,
Naveen Verma and James C. Sturm\thanks{This work was supported by the Semiconductor Research Corporation
(SRC), DARPA, Princeton Program in Plasma Science and Technology,
and Princeton University. \emph{(Corresponding author: Zhiwu Zheng)}}\thanks{The authors are with the Department of Electrical and Computer Engineering,
Princeton University, Princeton, New Jersey 08544, U.S.A. (e-mail:
zhiwuz@princeton.edu; hsin@princeton.edu; prakhark@princeton.edu;
wagner@princeton.edu; minjie@princeton.edu; nverma@princeton.edu;
sturm@princeton.edu).}}

\begin{document}
\bstctlcite{IEEEexample:BSTcontrol}
\maketitle
\begin{abstract}
Electrostatic actuators provide a promising approach to creating soft robotic sheets, due to their flexible form factor, modular integration, and fast response speed. 
However, their control requires kilo-Volt signals and understanding of complex dynamics resulting from force interactions by on-board and environmental effects. In this work, we demonstrate an untethered planar five-actuator piezoelectric robot powered by batteries and on-board high-voltage circuitry, and controlled through a wireless link. The scalable fabrication approach is based on bonding different functional layers on top of each other (steel foil substrate, actuators, flexible electronics).
The robot exhibits a range of controllable motions, including bidirectional crawling (up to
\textasciitilde{}0.6 cm/s), turning, and in-place rotation (at \textasciitilde{}1 degree/s). High-speed videos and control experiments show that the richness of the motion results from the interaction of an asymmetric mass distribution in the robot and the associated dependence of the dynamics on the driving frequency of the piezoelectrics. The robot's speed can reach 6 cm/s with specific payload distribution.
\end{abstract}

\section{INTRODUCTION}

Soft robots have gained much interest because of their ability to
achieve rich motions for moving around complex environments.
In particular, electrostatic soft robots made of piezoelectric actuators or dielectric elastomers can have small form factors \cite{Jafferis2019} and fast response speed \cite{Wu2019,Ji2019}. 

Untethered soft robots are essential for deployment and study in real-world applications.
While recent work has demonstrated untethered pneumatic \cite{T.2014,Duggan2019}
and microfluidic \cite{Wehner2016} soft robots, electrostatic
soft robots have posed challenges for integrating the drive and control electronics, because: (1) electrostatic actuators require high voltages, of hundreds or thousands of Volts, for strong actuation; (2) the voltage switching frequency needs to be fast, on the order of tens to hundreds of Hertz, to exploit the actuator's mechanical resonances; (3) the weight of the electronics significantly impacts the motion, and necessitates light weight to avoid suppressing the motion entirely. While recent research has led to lightweight untethered and fully-integrated soft robots made of electrostatic actuators requiring a few hundred Volts \cite{Jafferis2019,Ji2019,Liang2021}, as well as 
kilo-Volt soft robots that use buoyancy to reduce loading \cite{Li2017}, demonstrations of fully-untethered, wirelessly-controlled kilo-Volt soft robots moving on the ground with light-weight power circuits are limited. 

A primary challenge is that the dynamics of soft robots leading to controllable motions are challenging to
understand, due to substantial impact of forces on soft bodies arising from on-board and environmental components. This has motivated research on soft-robot motion, including modeling and control of linear and jumping movements \cite{Zheng2021,Zheng2022, Zheng2022a} as well as steering (turning), based on active differential
driving perpendicularly to the forward/backward
moving direction \cite{Vartholomeos2006,Pullin2012,Ho2013}, or adding electro-adhesive
frictional pads \cite{Liang2021,Chen2016}. Recent work has also explored motion arising from complex vibrational modes in specific structures, leading to frequency-controllable movements and rotations of an insect-scale soft robot \cite{Liang2019} and a miniature legged robot
\cite{Hariri2016,Dharmawan2017}. But studies extending to more generalizable structures have been limited. 

In this work, we address these challenges by developing a scalable and generalizable approach to fabricating untethered piezoelectric soft robots, and demonstrate this for a fully-wireless five-actuator robot. We develop miniaturized light-weight and modular power electronics on this kilo-volt robot, and explore a range of motions, including forward/backward crawling/turning and in-place rotation, controllable through the frequency of the actuation sequence and arising from the interaction of force profiles due to on-board components and vibrations in the linear robot structure. The robot is powered from on-board 3.7-V batteries and driven by on-board Bluetooth module, microcontroller, and custom power-electronics. The various frequency-dependent motion mechanisms originate from the intrinsic unbalanced weight distribution on the robot. The motion mechanisms are analyzed from experiments.

This paper has the following sections. Section \ref{sec:robot-structure} describes the fabrication approach and demonstrated robot structure. Section \ref{sec:forward-and-backward-crawling}
presents the robot's forward/backward movements, analyzes the forward motion of the robot as a representative example, and proposes its motion mechanism, further validated by control experiments.
Section \ref{sec:motion-mechanism}
demonstrates clockwise/counterclockwise rotations of the robot and analyzes the rotation mechanism.

\section{ROBOT STRUCTURE AND FABRICATION \label{sec:robot-structure}}
\begin{figure}
\subfloat[\label{fig:robot-schematics-side-separate}]{\begin{centering}
\includegraphics[width=0.9\columnwidth]{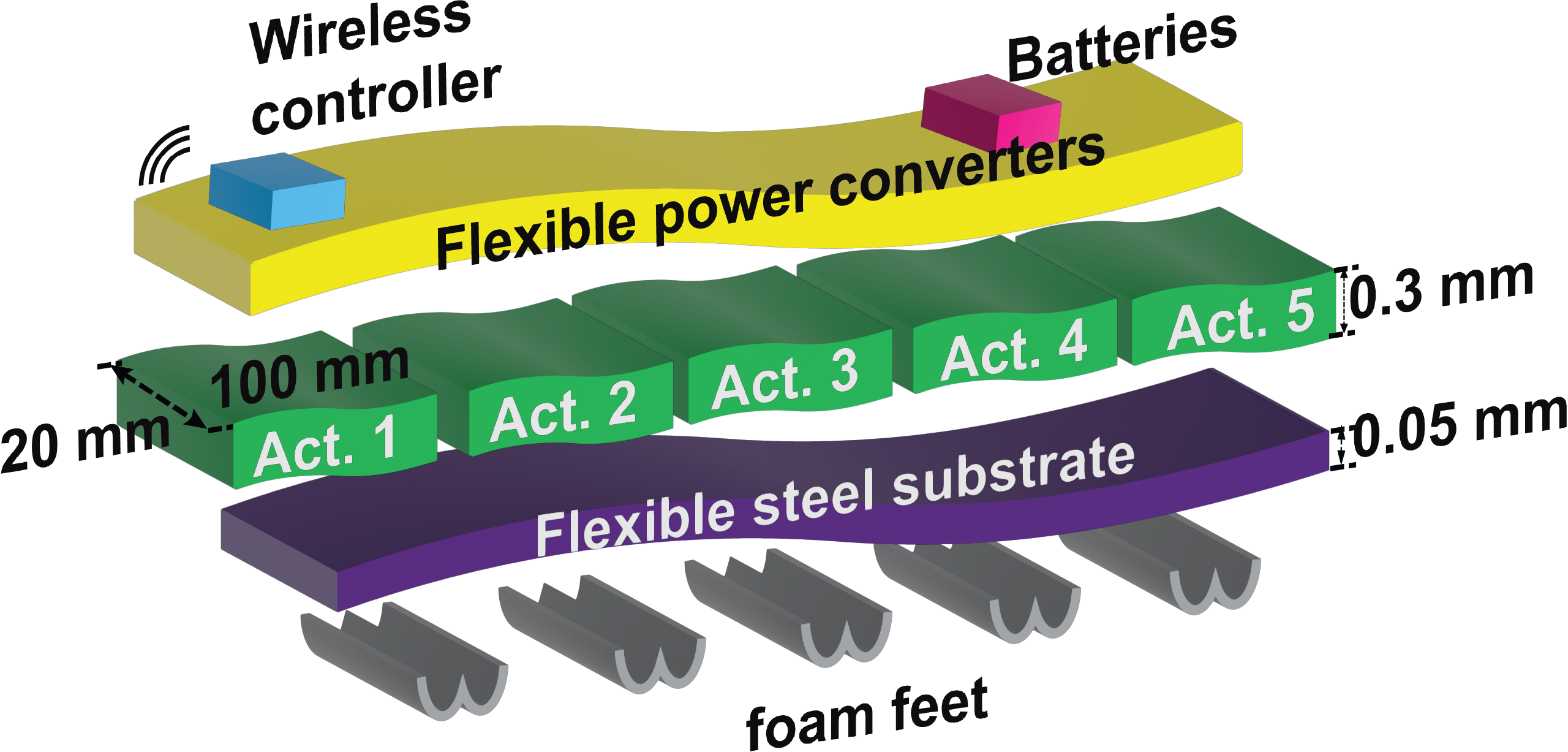}
\end{centering}
}

\subfloat[\label{fig:robot-schematics-side}]{\begin{centering}
\includegraphics[width=0.9\columnwidth]{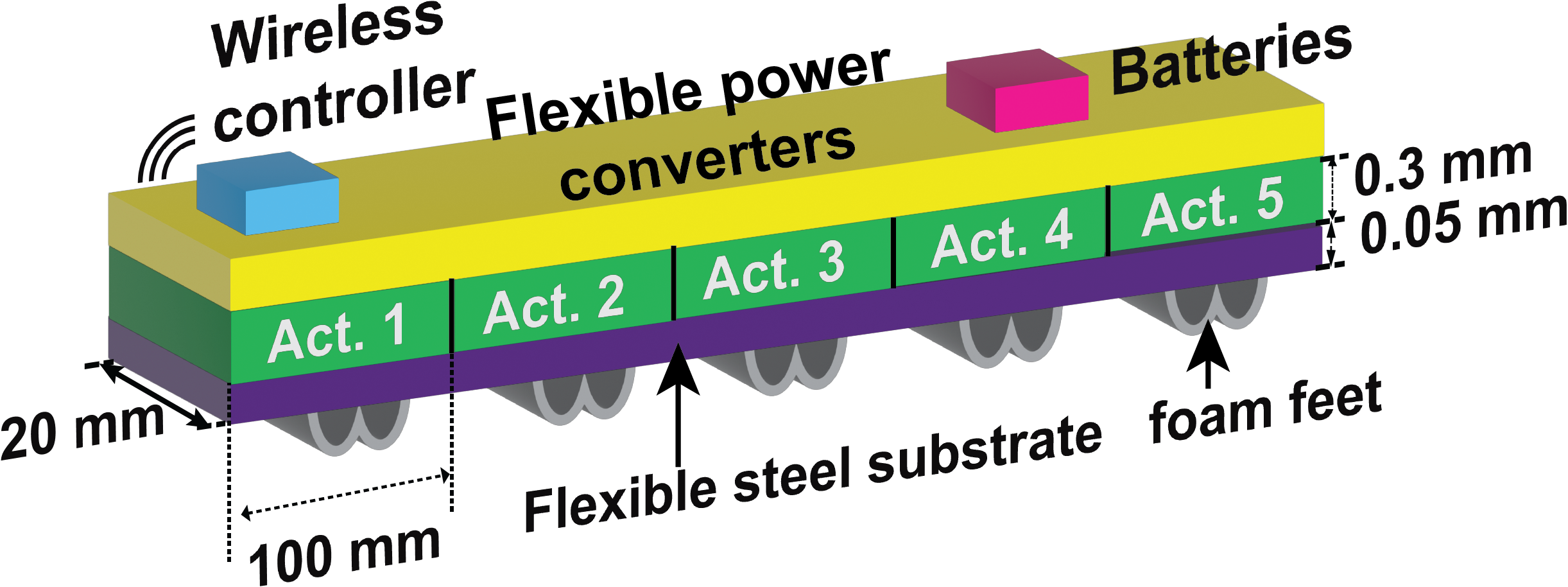}
\end{centering}
}

\subfloat[\label{fig:bending-mechanism}]{\begin{centering}
\includegraphics[width=0.9\columnwidth]{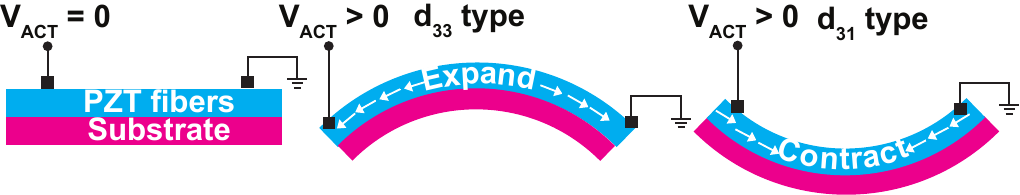}
\end{centering}
}

\caption{(a) Separated and (b) assembled views for
battery-powered wirelessly-controlled robot structure, 500 mm long
and 20 mm wide. Five ``B''-shape hollow foam feet are attached on
the bottom side, under the middle of each actuator. (c) Bending mechanism
of a single actuator made of PZT fibers bonded on steel foil substrate.
When voltage is applied, the $\text{d}_{33}$-type ($\text{d}_{31}$-type) PZT composite layer
tries to extend (contract), while the substrate does not. As a result,
the whole structure bends concave down (up). \label{fig:robot-schematics}}
\end{figure}

Figs. \ref{fig:robot-schematics-side-separate} and \ref{fig:robot-schematics-side} illustrate the fabrication approach and overall structure
of the demonstrated robot. Fabrication is hierarchical, based on separating the sub-fabrication of technologies required for different functionality (powering/control electronics, actuation, mechanical structuring) across different soft and flexible layers. This follows the scalable fabrication approach employed for commercial flat-panel displays, which today approach dimensions of up to 3 $\times$ 3 m \cite{Hendy2018,Montbach2018}.

The demonstrated robot comprises five 100-mm-long 300-\textmu m-thick
commercial piezoelectric composite units \cite{Smartmaterial}, which are
bonded to a 50-\textmu m-thick steel foil. The overall width is 20
mm. Five ``B''-shape hollow foam feet are bonded to the bottom side
of the robot. On top of the piezoelectric units, we developed and
connected a custom-designed flexible circuit board that implements
power converters, to convert battery-supplied 7.4 V into 300 V and
1500 V control signals for driving the actuators. 
The power converters utilize a new hybrid powering architecture which is not a focus of this paper and has been published elsewhere \cite{Cheng2022}. It features lightweight, high efficiency, high power density, and high modality compared with the state of the art. 
A Bluetooth module and microcontroller
\cite{dfrobot} are connected to the power converters for wireless
control.

Fig. \ref{fig:bending-mechanism} illustrates the bending mechanism
of a single actuator unit, which provides the basic movement of the robot. When voltage is applied,
the $\text{d}_{33}$-type ($\text{d}_{31}$-type) PZT layer tends to expand (contract), while the steel foil resists
due to its high stiffness. As a result, the actuator bends concave
down (up) instead. For this robot, only the middle actuator (Act. \#3) is $\text{d}_{33}$-type. Its applied voltage can go up to 1500 V. All the other actuators are $\text{d}_{31}$-type, on 300 V.

\begin{figure}
\begin{raggedright}
\begin{centering}
\subfloat[\label{fig:robot-picture-top}]{\begin{centering}
\includegraphics[width=1.\columnwidth]{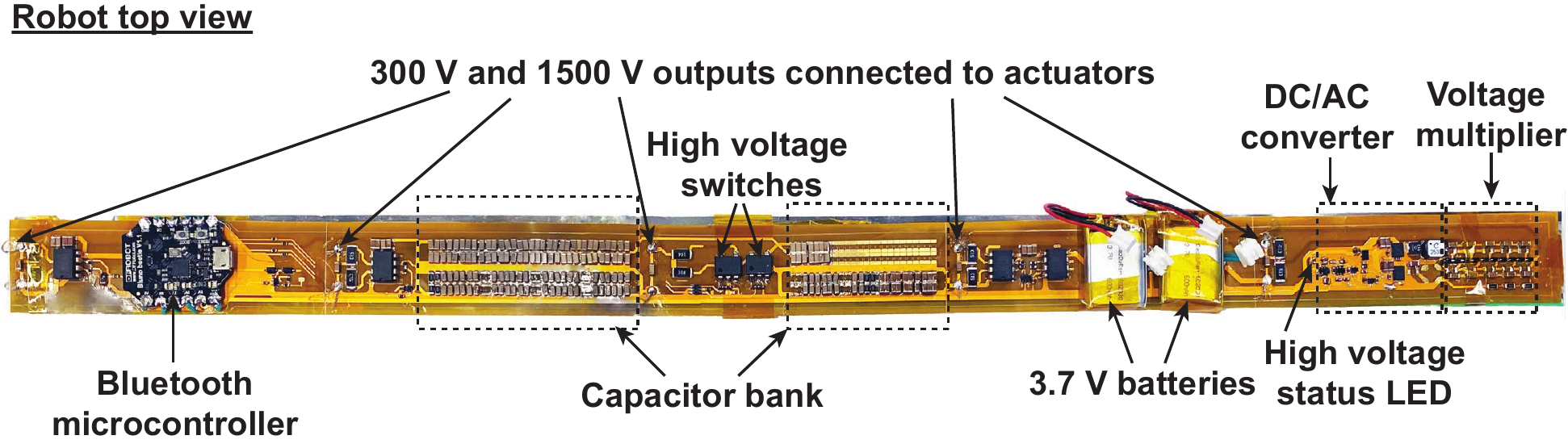}
\end{centering}
}\vfill{}
\end{centering}
\end{raggedright}

\begin{raggedright}
\begin{centering}
\subfloat[\label{fig:robot-picture-side}]{\begin{centering}
\includegraphics[width=1.\columnwidth]{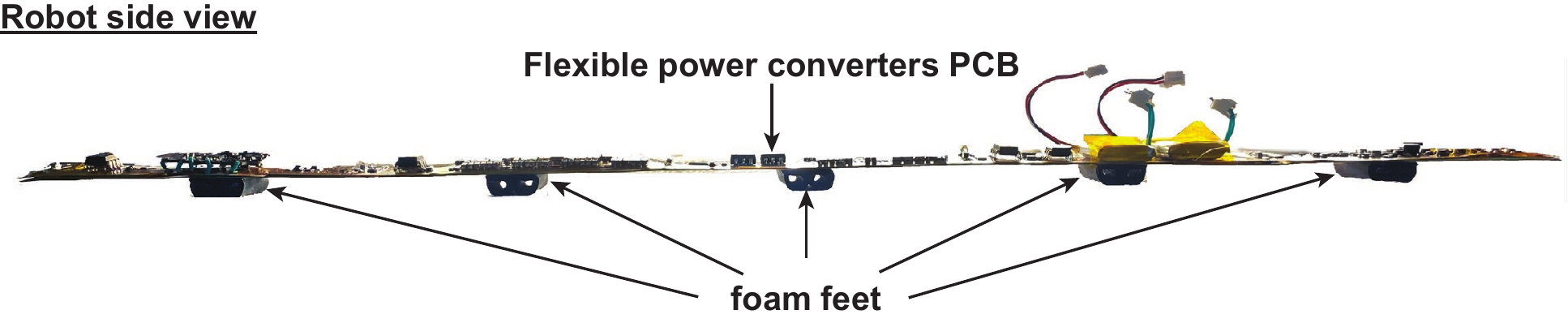}
\end{centering}
}\vfill{}
\end{centering}
\end{raggedright}
\caption{Pictures of the robot: (a) top view including power converters, Bluetooth module,
microcontroller, and batteries; (b) side view showing foam feet attached to the robot's bottom side. \label{fig:robot-pictures}}
\end{figure}

Fig. \ref{fig:robot-picture-top} shows the top view of the assembled robot.
The top layer is the power unit. The DC/AC converter and the multi-stage
voltage multipliers produce high-voltage rail signals (300 V and 1500 V), and the high-voltage switches turn on/off to drive high voltages to the actuators, which are connected through conductive vias on the flexible circuit board. The capacitor bank stabilizes
the high voltage output, and the microcontroller on-board is connected
wirelessly to a computer (not shown) for remote control. Fig. \ref{fig:robot-picture-side}
shows the side view, including the microcontroller on the left side,
the batteries on the right side, and the foam feet on the bottom.

\section{FORWARD AND BACKWARD CRAWLING \label{sec:forward-and-backward-crawling}}
The robot can move forward (rightward)/backward (leftward)
by controlling the frequency of the actuator-drive signal sequence. The actuators are driven by a two-phase control
sequence, where all the actuators are first turned on simultaneously and then
turned off simultaneously, with a duty cycle of 50\%. We note that while other control sequences are being explored (for which the possibilities are exceedingly numerous even for five actuators), our focus here is on how vibrational modes at different frequencies interact with robot force profiles to generate controllable motions. During our experiments, the robot's motion is recorded, and for
each run, we overlaid images of the robot over time onto a single
picture. For each frequency, the robot runs over 30 times to confirm the motion's stability and robustness.

\begin{figure}
\begin{centering}
\subfloat[\label{fig:forward}]{\begin{centering}
\includegraphics[width=0.85\columnwidth]{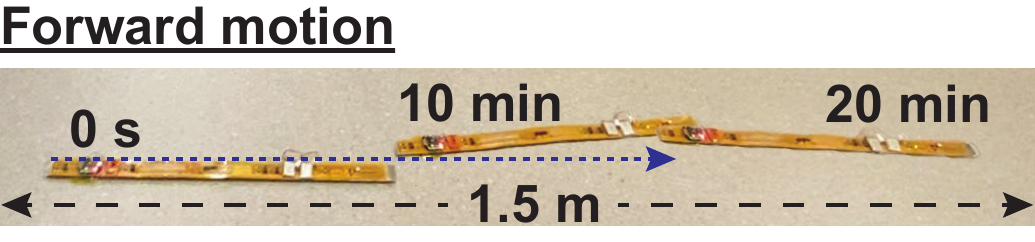}
\end{centering}
}

\subfloat[\label{fig:backward-anticlockwise}]{\begin{centering}
\includegraphics[width=0.85\columnwidth]{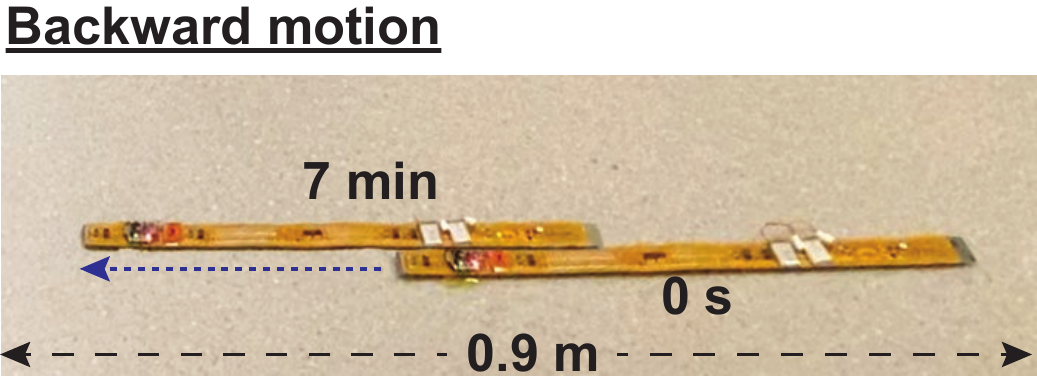}
\end{centering}
}

\caption{Snapshots of the robot's frequency-controlled lateral movement over
time: (a) forward motion, when driven at 13 Hz; (b) backward motion, when driven at
at 14 Hz.\label{fig:lateral-motion-pictures}}
\end{centering}
\end{figure}

Fig. \ref{fig:lateral-motion-pictures} demonstrate the robot's lateral motions. For example, the robot can move forward (Fig. \ref{fig:forward}) and backward (Fig. \ref{fig:backward-anticlockwise}), all by frequency modulation only.

\subsection{Motion characterization \label{sec:motion-characterization}}

Each frequency and each kind of motion contain more than 30 trails, and no noticeable variance in speed and direction is observed. With the carried batteries, the robot can run continuously for 40 minutes before the batteries run out.

\begin{figure}
\centering
\includegraphics[width=0.7\columnwidth]{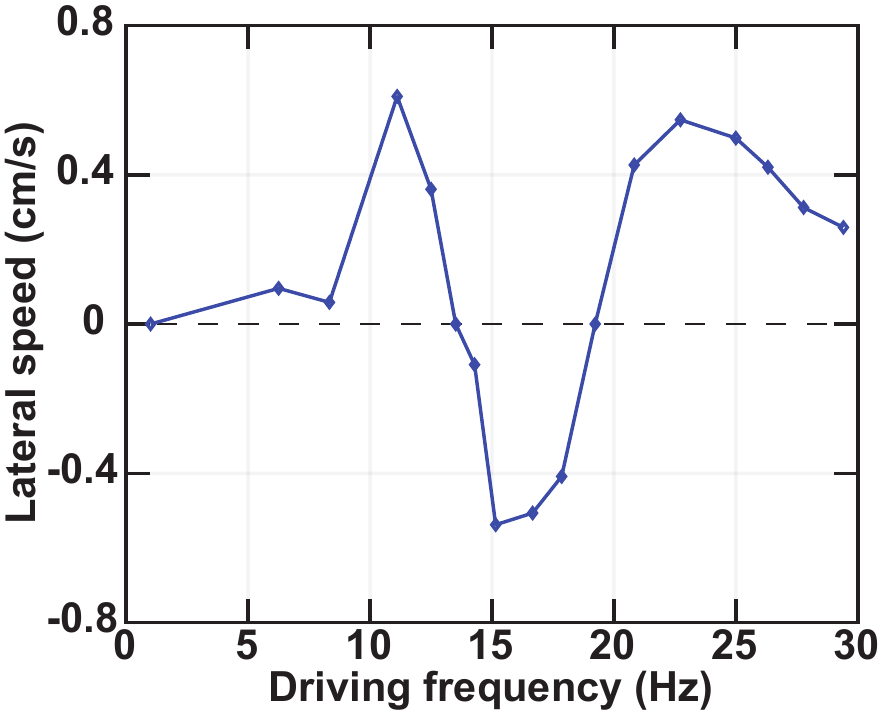}
\caption{Frequency dependence of the lateral speed. Positive numbers mean that the robot moves forward.
\label{fig:robot-speed-frequency}}
\end{figure}
Fig. \ref{fig:robot-speed-frequency} shows the frequency dependency of the forward and backward motions. The robot moves forward at low frequencies, up to 0.6 cm/s at 11 Hz; has no lateral motion at 14 Hz; then moves backward
as frequency increases, up to -0.5 cm/s at 15 Hz; has no
lateral motion again at 19 Hz; and finally moves forward again, up to 0.5 cm/s at 23 Hz.

\begin{figure}
\begin{centering}
\subfloat[\label{fig:forward-time-domain}]{\begin{centering}
\includegraphics[width=0.45\columnwidth]{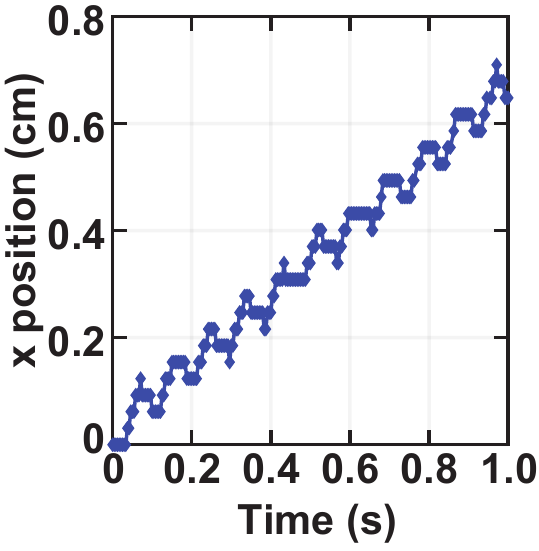}
\end{centering}
}\vfill{}
\subfloat[\label{fig:forward-side}]{\begin{centering}
\includegraphics[width=0.8\columnwidth]{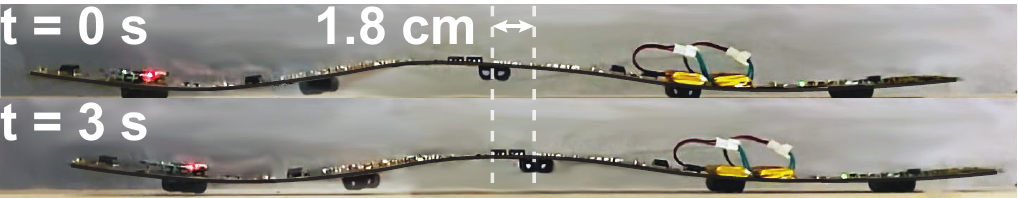}
\end{centering}
}

\caption{Motion examples in the time domain for: (a) lateral forward (rightward)
movement, when driven at 11 Hz; 
(b) side-view pictures for the forward motion (along x direction) over 3 seconds.
\label{fig:motion-time-domain}}
\end{centering}
\end{figure}

Fig. \ref{fig:motion-time-domain} plots examples of robot forward
motion vs. time. At 11 Hz, the
robot moves forward at 0.7 cm/second (Fig. \ref{fig:forward-time-domain}).
Side view pictures
(Fig. \ref{fig:forward-side}) show the forward motion by 1.8 cm within
3 seconds.

\subsection{Motion mechanism: weight asymmetry and frequency-dependent movements \label{sec:motion-mechanism}}

Modeling of this robot's motion is very challenging, as it includes dynamics (e.g. resonances) over a continuous free-moving body, time-varying external interactions (such as gravity and soft contact with the ground), and irregular weight distribution. As quantitative analytic modeling efforts are ongoing, this paper focuses on experimental illustration, analysis, and validation of the motion mechanism. As previously mentioned in Sec. \ref{sec:motion-characterization}, the motion behaviors are stable, tested over more than 30 trials for each motion.

When the robot is driven at high frequencies (> 1 Hz), its dynamics
play an important role in the motions observed. The bending actuators generate vertical
momentum and excitation waves, which travel along the robot's length. These not only cause individual segments to rise off the ground, but can make the entire robot jump off the ground.

\begin{figure}
\begin{centering}
\includegraphics[width=0.8\columnwidth]{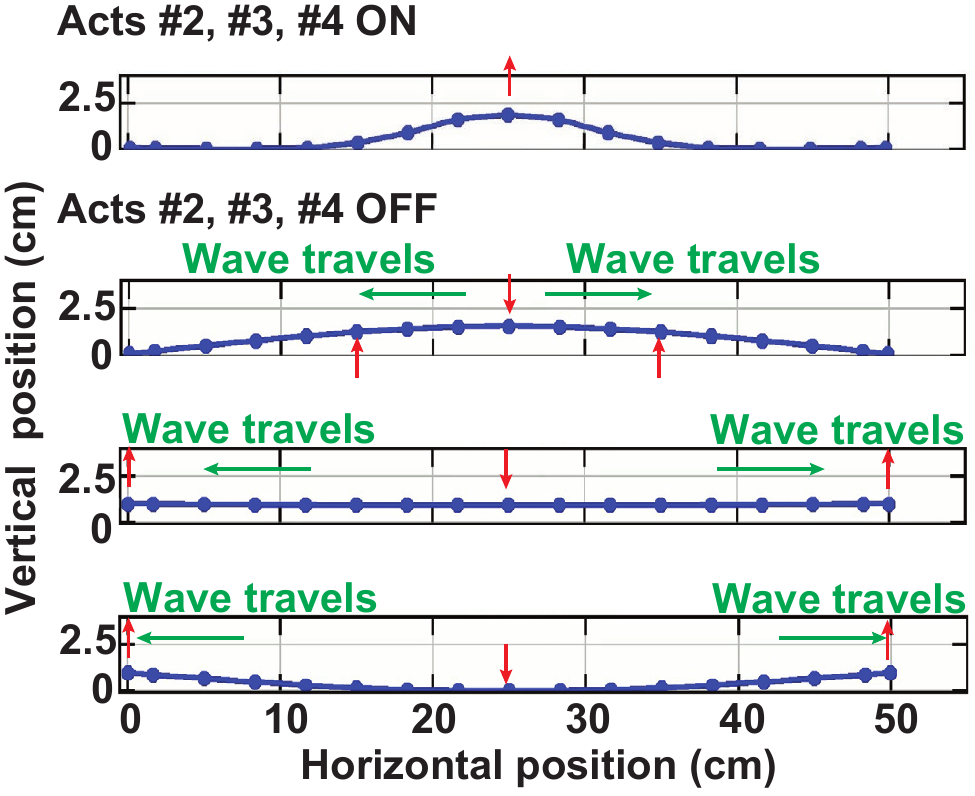}
\caption{Traveling waves propagating along the robot. Snapshots of the robot's shape over time are simulated and plotted. Voltages of actuators \#2, \#3, and \#4 are turned on and off. Traveling waves transmit from the midsection to the ends, which ultimately lift up as a result.
\label{fig:traveling-waves}}
\end{centering}
\end{figure}

Fig. \ref{fig:traveling-waves} shows PyBullet simulations of this effect, based on an integrated and experimentally-verified model of the actuators and overall robot \cite{Zheng2022}. As seen, starting from a position where the middle three actuators are bent and then unbent, an outward travelling wave is generated along the robot length, with a period where the entire robot is suspended off the ground.

Although the intrinsic waves are symmetric, if there is weight asymmetry along the length of the robot, the shape of the robot and the instances when it contacts the ground will be correspondingly affected. As an example, one end can contact the ground earlier, thereby generating lateral motion through ground-force friction, as the robot segments proceed to expand/contract during unbending/bending.

\begin{figure}
\begin{centering}
\includegraphics[width=0.8\columnwidth]{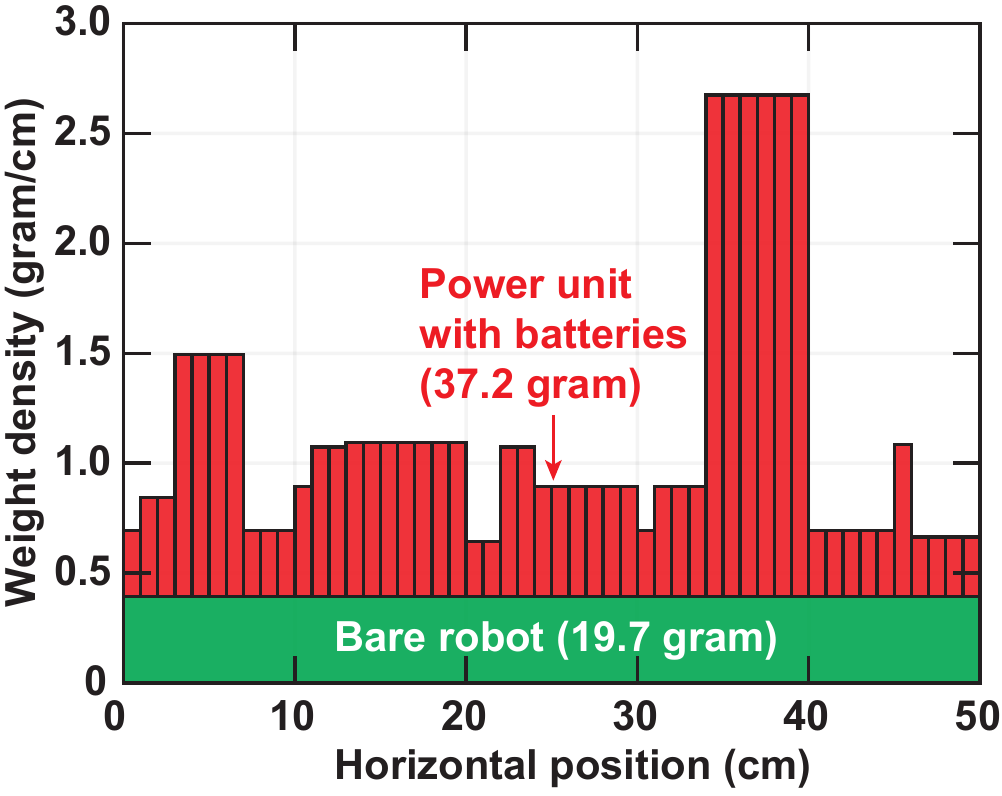}
\caption{Non-uniform weight distribution of the robot, including a bare robot
(actuators, substrate, and foam feet) (green) and the electronics on flexible PCB, including the microcontroller and batteries (red). \label{fig:weight-distribution}}
\end{centering}
\end{figure}

Fig. \ref{fig:weight-distribution} plots 1-D weight distribution
along the length of the robot. The distribution includes the 20-gram
bare robot (including the piezoelectric composites and steel
foil substrate with the five foam feet) and a 37-gram electronics
unit (microcontroller, power converters, batteries, etc.). The distribution shown in the plot is estimated from measurements of the bare robot, and the discrete
electrical components. Note that the scattered foam feet are very light, so their weights are ignored. While for a bare robot, the weight distribution is uniform over the length of the robot, it is highly non-uniform due to the electronics, with the right side having the highest density (2.7 gram/cm) at the location of the batteries and the left side having the highest density (1.5 gram/cm) at the location of the microcontroller. There are opportunities to further optimize the weight distribution on the robot to enhance the target motion.

\begin{figure}
\begin{centering}
\includegraphics[width=1.0\columnwidth]{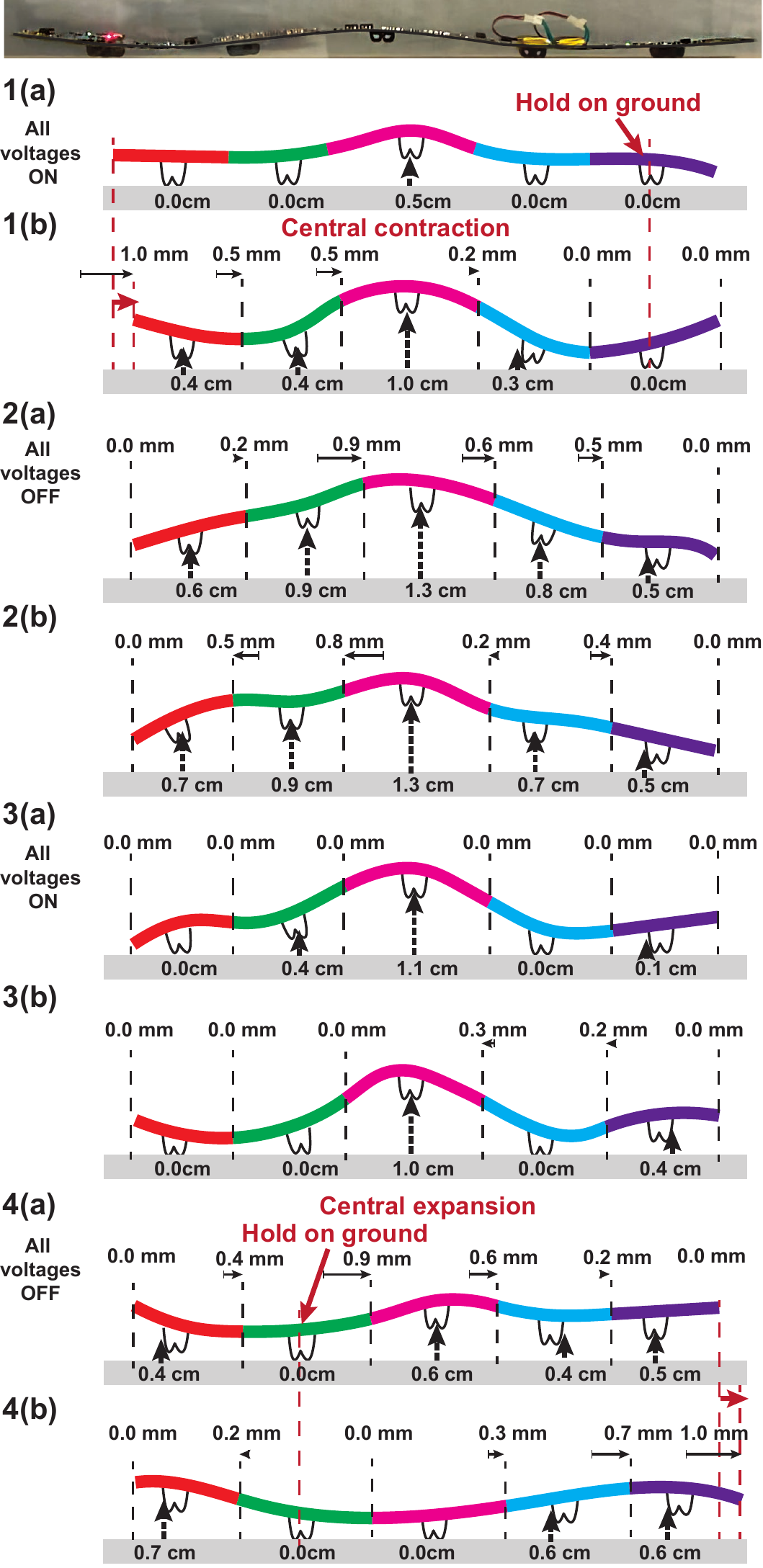}
\end{centering}
\caption{Mechanism of robot jumping to the right shown by a series of high-speed camera images (redrawn schematically for clarity). All five actuators are simultaneously turned on and off at 11 Hz. The elapsed time is 23 ms from one subfigure to the next. The forward movement (defined as to the right) is caused by "inchworm motion": the robot contracts its length while the right foot is on the ground and then extends while the left foot is on the ground. The asymmetry results from asymmetric weight distribution of the robot. Vertical arrows and numbers show the height from the ground for each foam foot, and horizontal arrows and numbers show lateral movement compared with the previous time step.
\label{fig:forward-schematics}}
\end{figure}

Next, we analyze how the weight distribution gives rise to motion, using the robot's forward motion (at 11 Hz) as an example. Fig.~\ref{fig:forward-schematics} illustrates the robot's shapes at different instances of the control-signal sequence (taken from high-speed cameras), with annotations along the top showing the experimentally-measured lateral movement at either end of each actuator and annotations at the rubber feet showing the experimentally-measured height at the corresponding location. The following steps over time are observed:
\begin{itemize}
\item Step 1: All the voltages are turned on (as in Fig. \ref{fig:forward-schematics}-
1(a),1(b)). Vertical momentum is generated, and the robot jumps
off the ground while its midsection is contracting. The left
part of the robot lifts off the ground first (Fig. \ref{fig:forward-schematics}-1(b)), while the right-most foam foot remains on the ground. With the right end contacting the ground and the midsection contracting, the left end moves rightward. 
\item Step 2: The voltages are turned off. The robot continues to rise due
to the generated momentum. Because the robot becomes flat, it
lifts off the ground (Fig. \ref{fig:forward-schematics}-2(a),2(b)).
\item Step 3: All the voltages are turned back on. However, since the robot is
still in the air, further vertical momentum is not generated without counter force from the ground. So the robot continues to fall, finally landing on the ground
(Fig. \ref{fig:forward-schematics}-3(b)). However, because the robot is contracting,
its midsection is relatively higher, and only the left and/or
the right ends land. At this point, more of the left end (first
and second foam feet) is on the ground than the right end (4th foam
foot). 
\item Step 4: The voltages are turned back off, causing the robot to extend and flatten. This again tends to lift it off the ground. As more
of the left end is on the ground, all foam feet except the 2nd left
one will lift off. With the second left foot remaining on the ground while the robot
extends, overall rightward movement is observed.
\end{itemize}

Over the course of the four steps, corresponding to two periods of the actuator driving signals, overall rightward motion of the robot is generated. We note that this motion is non-linear, with one cycle of motion two driving cycles.

Driving the robot at different frequencies will cause different
vibration modes. The vibrations determine which side of the robot
contacts the ground at different instances of contraction/expansion, and therefore determine the direction of the lateral movement.

\subsection{Experimental validation of motion mechanism with a "bare" robot \label{sec:experimental-validation}}

Having analyzed the robot's forward motion
as a representative example, to understand the frequency and weight-distribution
dependence of movement, here we validate the impact of these two
mechanisms respectively in two control experiments. We first examine a case where the weight distribution is uniform, and a case with simple non-uniformity. In both cases, the robot's actuation frequency is swept from 1 Hz to 30 Hz.

\begin{figure}
\centering
\subfloat[\label{fig:symmetric-setup}]{\begin{centering}
\includegraphics[width=0.45\columnwidth]{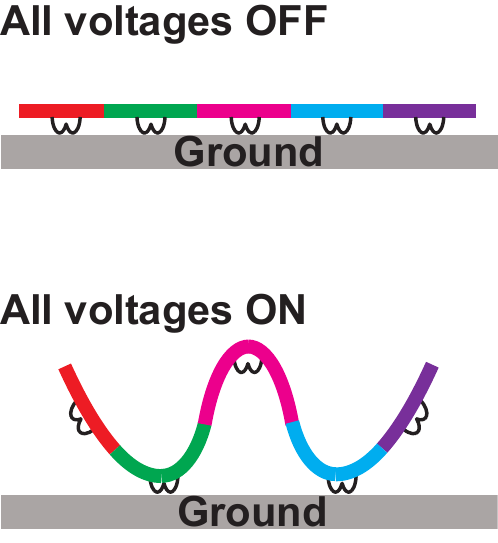}
\end{centering}
}\subfloat[\label{fig:symmetric-horizontal-motion}]{\begin{centering}
\includegraphics[width=0.45\columnwidth]{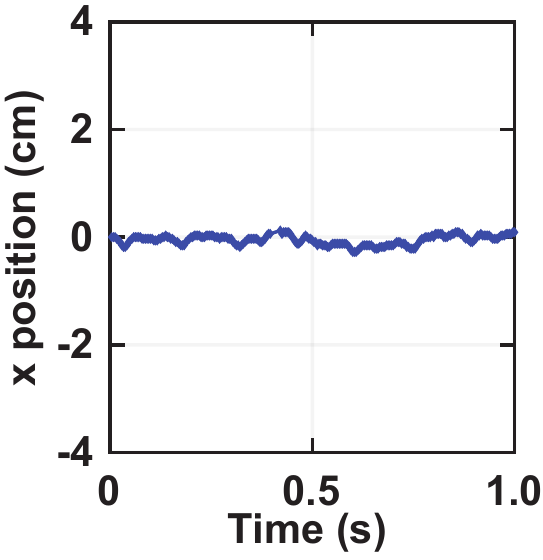}
\end{centering}
}

\subfloat[\label{fig:sym-speed-frequency}]{\begin{centering}
\includegraphics[width=0.55\columnwidth]{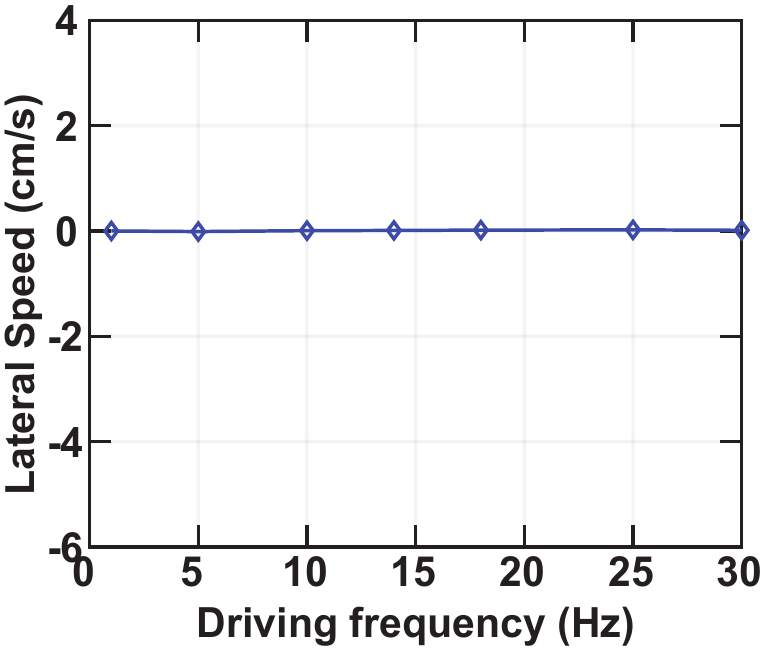}
\end{centering}
}
\caption{(a) Robot setup and (b) the horizontal movement over time for a uniformly-loaded
setup (bare robot, only comprising the actuators and feet).(c) Frequency dependence of the robot's lateral speed, with the uniform setup in Fig. \ref{fig:symmetric-setup}. \label{fig:symmetric-experiment}}
\end{figure}

Fig. \ref{fig:symmetric-setup} shows the experimental setup consisting of just the actuators and feet, but no electronics components, for uniform weight distribution. The high voltages required for actuator control are supplied by external circuitry connected by ultra-thin wires, which exhibit minimal force on the robot. The voltage signals are applied at different frequencies with 50\% duty cycle. We refer to this as a ``bare robot''. 

Unlike the fully-loaded robot case, the bare robot is observed to exhibit no lateral motion at any frequency (Fig. \ref{fig:sym-speed-frequency}). Fig. \ref{fig:symmetric-horizontal-motion} shows an example, with
lateral position of the robot vs. time, no net motion is observed.

\begin{figure}
\centering
\subfloat[\label{fig:asym-weight-setup}]{\begin{centering}
\includegraphics[width=0.8\columnwidth]{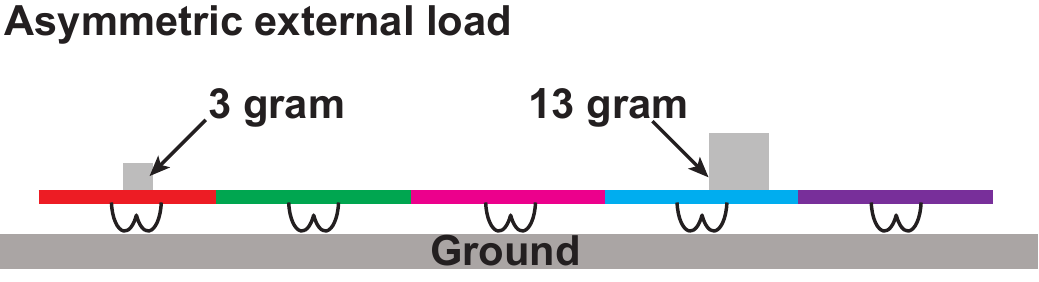}
\end{centering}
}

\subfloat[\label{fig:asym-forward}]{\begin{centering}
\includegraphics[width=0.35\columnwidth]{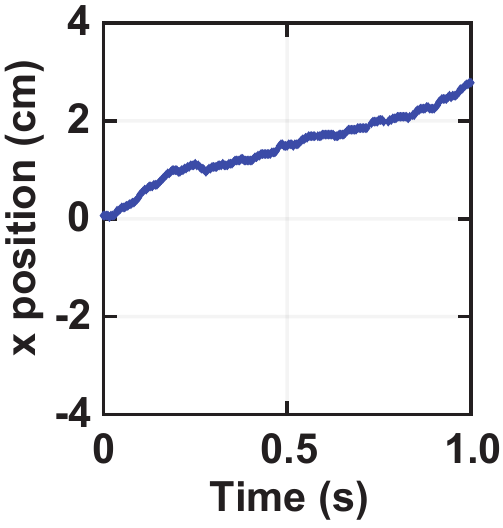}
\end{centering}
}\subfloat[\label{fig:asym-backward}]{\begin{centering}
\includegraphics[width=0.35\columnwidth]{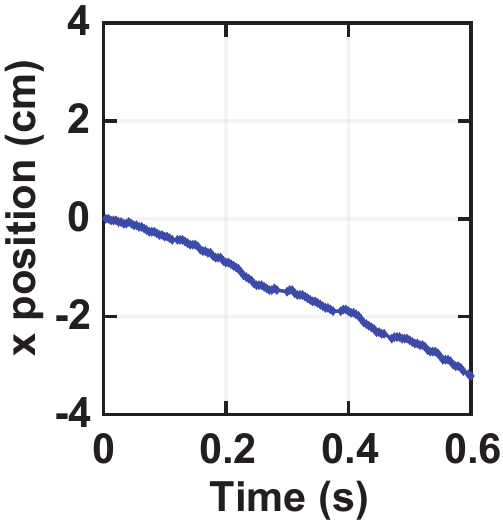}
\end{centering}
}

\subfloat[\label{fig:asym-speed-frequency}]{\begin{centering}
\includegraphics[width=0.55\columnwidth]{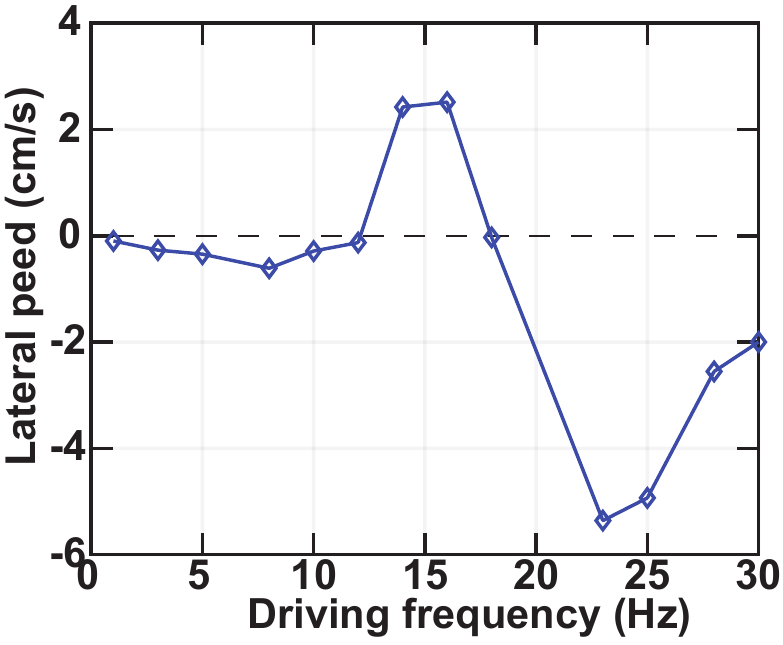}
\end{centering}
}
\caption{Robot with non-uniform loading. (a) Robot setup with a 3-gram weight on
the middle of actuator \#1 and 13-gram weight on actuator \#4. (b) Robot forward motion of 2.5 cm/s, when driven at 16 Hz. (c) Robot backward motion
of -5.4 cm/s, when driven at 23 Hz. (d) Frequency dependence of the robot's lateral speed, with the non-uniform setup in Fig. \ref{fig:asym-weight-setup}.
\label{fig:asym-setup}}
\end{figure}

Fig. \ref{fig:asym-setup} shows the experimental setup where a simple
weight non-uniformity is added, which we observe to cause frequency-dependent lateral motion. A 3-gram weight is attached at the midpoint of the robot's actuator \#1, and another
13-gram weight is attached at the end of actuator \#4 (Fig. \ref{fig:asym-weight-setup}).
The same voltage sequence is applied while sweeping the driving frequency. Here, the robot's lateral speed and direction change significantly with
frequency. Fig. \ref{fig:asym-speed-frequency} shows that: as frequency increases, the robot moves backward, up to -0.61 cm/s at 8 Hz; changes direction to move forward, up to 2.5 cm/s at 16 Hz; and finally moves backward again, up to -5.4 cm/s at 23 Hz. Figs. \ref{fig:asym-forward} and \ref{fig:asym-backward} plot the robot's lateral positions versus time for two different driving
frequencies. The robot moves forward by 2.5 cm/s when driven at 16 Hz
and backward by -5.4 cm/s when driven at 23 Hz. 

The analysis in this paper has focused on forward/reverse motion and its dependence on the driving frequency, with a key requirement being non-uniformity in the distribution of weight along the robot's length. Exploration of the observed frequency-dependent rotation is still on-going, with preliminary understanding suggesting that this arises due to weight-distribution non-uniformity across the width of the robot.
\section{BIDIRECTIONAL TURNING \label{sec:birectional-rotation}}
\begin{figure}
\centering
\subfloat[\label{fig:in-place-clockwise}]{\begin{centering}
\includegraphics[height = 1.5in]{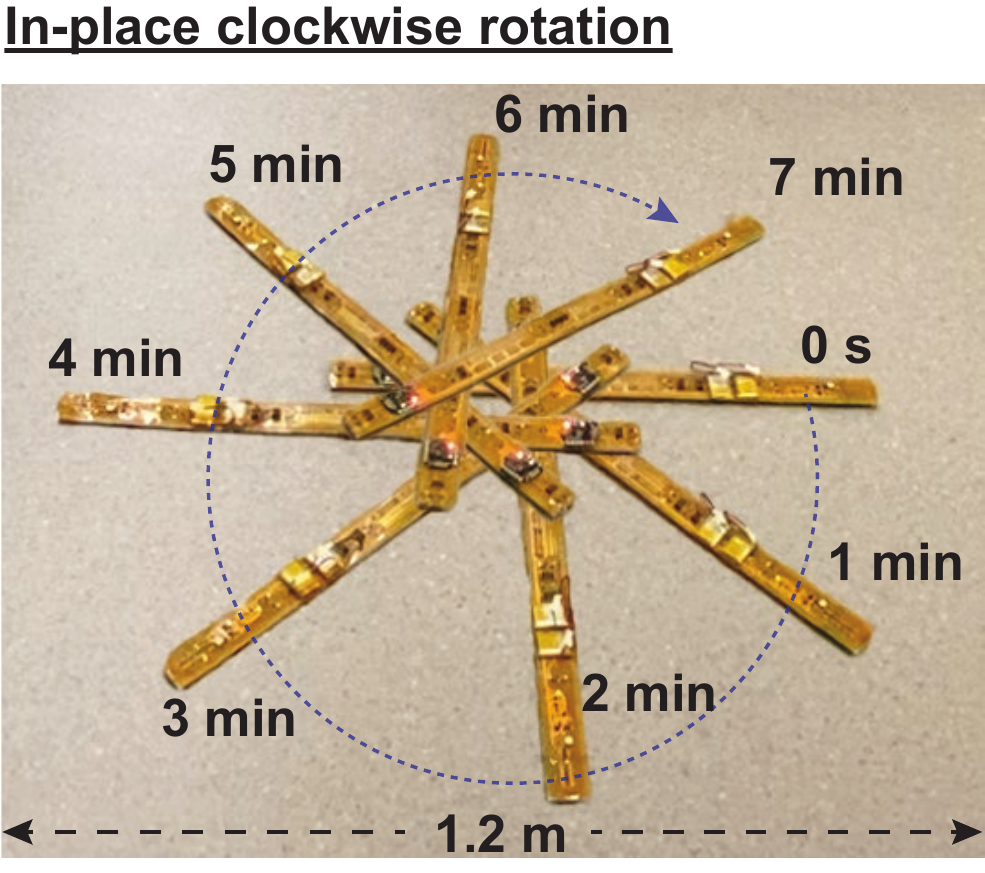}
\end{centering}
}
\subfloat[\label{fig:forward-clockwise}]{\begin{centering}
\includegraphics[height = 1.5in]{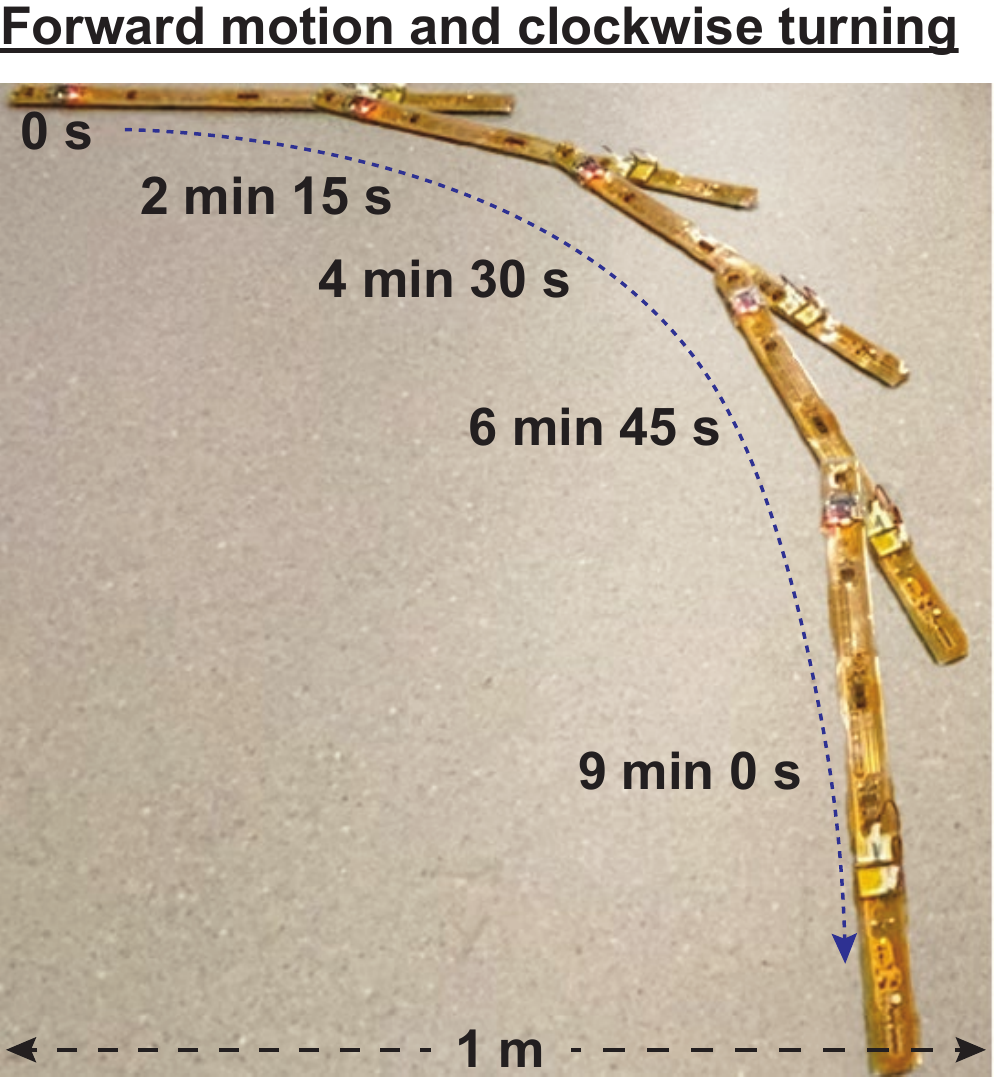}
\end{centering}
}
\caption{Examples of bidirectional turning: (a) in-place clockwise turning, when driven at 23 Hz; (b) Forward motion while turning counterclockwise, when driven at 21 Hz.
\label{fig:bidirectional-turning}}
\end{figure}

The robot also can turn counterclockwise/clockwise,
by controlling the frequency of the actuator-drive signal sequence. Fig. \ref{fig:bidirectional-turning} demonstrate the robot's rotations. For example, the robot can rotate clockwise in-place
(Fig. \ref{fig:in-place-clockwise}) and it also can move forward while turning
clockwise (Fig. \ref{fig:forward-clockwise}), all by frequency modulation only.
\begin{figure}
\begin{centering}
\includegraphics[width=0.4\columnwidth]{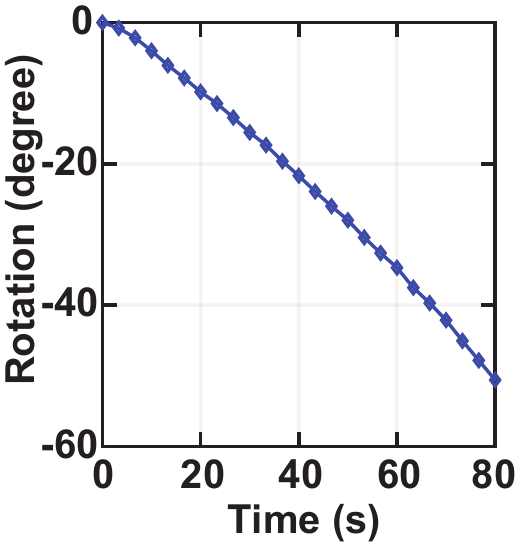}
\caption{Time domain experiment data for the example of in-place clockwise rotation (driven at 23 Hz) as in Fig. \ref{fig:in-place-clockwise}.
\label{fig:rotation-time-domain}}
\end{centering}
\end{figure}

Fig. \ref{fig:rotation-time-domain} plots time-domain experiment data for the example of in-place rotation as in Fig. \ref{fig:in-place-clockwise}
driven at 23 Hz. The robot rotates in-place by 50 degrees clockwise in 80
seconds.

\begin{figure}
\centering
\includegraphics[width=0.7\columnwidth]{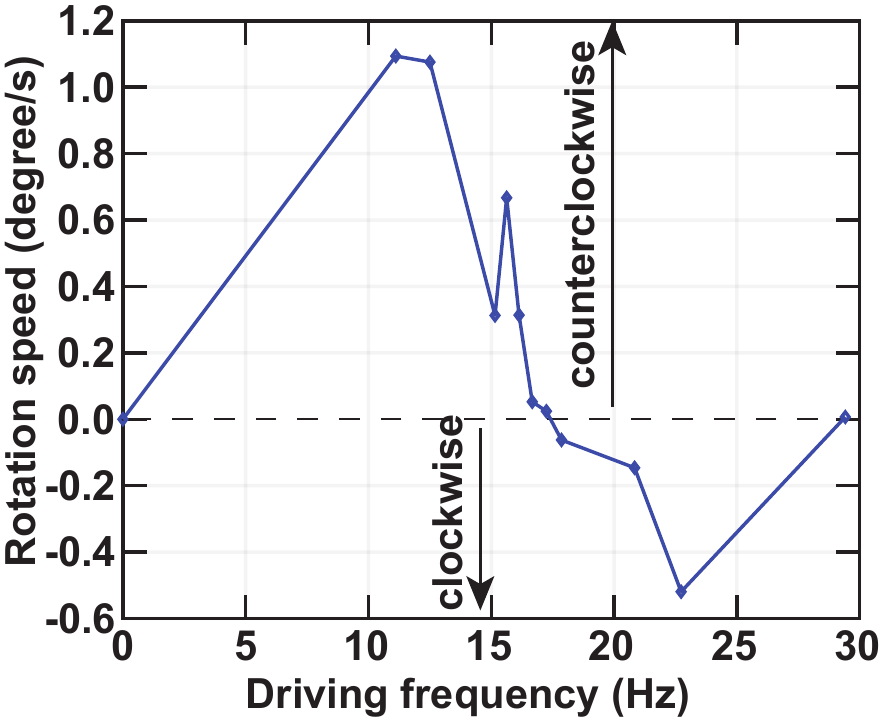}
\caption{Frequency dependence of rotation speed.
\label{fig:robot-rotation-frequency}}
\end{figure}

Fig. \ref{fig:robot-rotation-frequency} shows the frequency dependency of the rotation. The robot turns counterclockwise
at low frequencies, up to 1.1 degree/second at 11 Hz; has no
rotation at 17 Hz; then rotates clockwise as frequency further
increases, up to -0.5 degree/second at 23 Hz.

The mechanism analysis can be extended from Sec. \ref{sec:motion-mechanism}. The robot forward/backward motion is due to weight asymmetry at the midpoint perpendicular to the length of the robot. Similarly, the rotation is caused by the weight asymmetry at the midpoint along the length direction. We can understand this by dividing the robot into the front and back halves. For example, at a specific frequency where the front half tends to move forward (rightward), while the back half wants to move backward (leftward) due to asymmetry with the front half. The whole robot will rotate anticlockwise from the top view. The rotational direction and speed can be controlled by driven frequency as demonstrated in the experiment.

\section{CONCLUSION}

This work develops a platform for wirelessly-controlled battery-powered untethered
planar soft robots made of piezoelectric actuators, and demonstrates a prototype comprised of five piezoelectric actuators bonded to a steel substrate.
Due to both weight non-uniformity along the length of the robot and drive-frequency-dependent vibrations arising in the robot, it can move forwards or backwards, controllable through the drive frequency, at up to $\sim$ 0.6 cm/s. Crucially, the robot's left/right ends contact the ground during different times during its expansion/contraction cycles. This paper focuses on experimental exploration and validation of the motion mechanism. The motion mechanism was experimentally
validated by testing "bare" robots with both uniform and asymmetric mass distributions. With a specific payload distribution, the robot's speed can reach 6 cm/s.
The robot can also rotate clockwise or counterclockwise in the range of -0.5 $\sim$ 1.1 degree/s. The rotation mechanism is analyzed qualitatively. Future work includes theoretical and quantitative understanding and modeling of the motion mechanisms and optimization of the movement speed by weight distribution. 

\bibliographystyle{IEEEtran}
\bibliography{IEEEabrv, repeat_name, mybib}

\end{document}